\documentclass[sigconf, nonacm]{acmart}

\usepackage{amsmath}
\usepackage{graphicx}
\usepackage{enumitem}
\usepackage{algorithm}
\usepackage{algorithmic}
\usepackage{multirow}
\usepackage{hyperref}
\usepackage{tikz}
\usepackage{subcaption}
\usetikzlibrary{positioning,arrows.meta}

\newcommand\vldbyear{2026}
\newcommand\vldbworkshop{DASHSys: Systems for Data-centric Agents with Human-in-the-loop}
\newcommand\vldbauthors{\authors}
\newcommand\vldbtitle{\shorttitle}

\newcommand\vldbpagestyle{plain}

\begin{document}

\title{ASMR: Agentic Schema Generation for Ship Maintenance Report Writing}
\author{Sohrab Namazi Nia}
\affiliation{%
  \institution{New Jersey Institute of Technology}
  \city{Newark}
  \state{New Jersey}
  \country{USA}
}
\email{sn773@njit.edu}

\author{Amogh Dalal}
\affiliation{%
  \institution{New Jersey Institute of Technology}
  \city{Newark}
  \state{New Jersey}
  \country{USA}
}
\email{asd33@njit.edu}

\author{Ning Sa}
\affiliation{%
  \institution{Rensselaer Polytechnic Institute}
  \city{Troy}
  \state{New York}
  \country{USA}
}
\email{san2@rpi.edu}

\author{Peter Ly}
\affiliation{%
  \institution{Rensselaer Polytechnic Institute}
  \city{Troy}
  \state{New York}
  \country{USA}
}
\email{lyp2@rpi.edu}

\author{Marti Zentmaier}
\affiliation{%
  \institution{Boston Fusion Corporation}
  \city{Lexington}
  \state{Massachusetts}
  \country{USA}
}
\email{marti.zentmaier@bostonfusion.com}

\author{Tomek Strzalkowski}
\affiliation{%
  \institution{Rensselaer Polytechnic Institute}
  \city{Troy}
  \state{New York}
  \country{USA}
}
\email{tomek@rpi.edu}

\author{Jay Miller}
\affiliation{%
  \institution{Boston Fusion Corporation}
  \city{Lexington}
  \state{Massachusetts}
  \country{USA}
}
\email{jay.miller@bostonfusion.com}

\author{Rishi Singh}
\affiliation{%
  \institution{Boston Fusion Corporation}
  \city{Lexington}
  \state{Massachusetts}
  \country{USA}
}
\email{rishi.singh@bostonfusion.com}

\author{Senjuti Basu Roy}
\affiliation{%
  \institution{New Jersey Institute of Technology}
  \city{Newark}
  \state{New Jersey}
  \country{USA}
}
\email{senjutib@njit.edu}

\begin{abstract}
%Operational and maintenance reporting systems in ships often rely on forms containing substantial unstructured textual narratives describing equipment conditions, hazards, mitigation actions, system failures, and operational observations. While these narratives contain valuable operational knowledge, they are frequently inconsistent, incomplete, and difficult to analyze at scale. A key challenge is that different form categories require different information requirements and reporting best practices, many of which are not explicitly documented and instead reside in historical reports and expert knowledge.

In this paper, we study the \emph{automatic schema generation problem}: given a collection of historical ship maintenance and operational reports across multiple form categories, automatically discover compact and informative schemas that capture the essential information requirements of each report type.  To address this challenge, we propose {\bf ASMR}, a modular agentic framework consisting of two specialized agents. A \emph{Field Generation Agent} extracts semantic concepts from historical narratives and generates candidate schema fields through adaptive multi-granularity clustering, while a \emph{Structural Optimizer Agent} employs reinforcement learning to identify compact, informative, and non-redundant schema representations. The resulting schemas can guide report authors toward producing more complete, consistent, and actionable reports. Preliminary results demonstrate the promise of the proposed approach and highlight several open research challenges at the intersection of data management, agentic AI, and human-centered AI.

\end{abstract}

\maketitle

%%% VLDB block start %%%
\pagestyle{\vldbpagestyle}
\begingroup\small\noindent\raggedright\textbf{VLDB Workshop Reference Format:}\\
\vldbauthors. \vldbtitle. VLDB \vldbyear\ Workshop: \vldbworkshop.
\endgroup

\begingroup
\renewcommand\thefootnote{}\footnote{\noindent
This work is licensed under the Creative Commons BY-NC-ND 4.0 International License.
}\addtocounter{footnote}{-1}
\endgroup

\section{Introduction}
Many ship operational and maintenance reporting workflows rely on forms that contain substantial unstructured textual narratives. Free-form text provides flexibility to describe critical operational information such as equipment conditions, hazards, mitigation actions, system failures, preservation activities, maintenance observations, and operational impacts. However, across different personnel with varying levels of expertise, this same flexibility can result in inconsistent structure that makes it difficult to extract actionable insights from  reports at scale. In particular, required information that goes unrecorded by personnel can be impossible to recover, which is a problem made worse in large enterprises where report writers are far removed from downstream report consumers. Instead of burdening personnel with more training that scales poorly across large workforces, we imagine real-time writing assistance that provides topic-specific information requirements to personnel during the writing process. This would better optimize tradeoffs between flexibility and structure provided by free-text inputs while supporting downstream workflows.

To support such a capability, we formulate the \emph{schema generation problem}: given a collection of historical reports, automatically discover schemas consisting of compact sets of information fields that capture essential concepts and report information requirements; additionally, as the relevant concepts and information requirements for a report will change with its general topic, each schema should correspond to a specific report topic. A schema field represents a recurring semantic aspect of the reports, such as \emph{Condition}, \emph{Hazard}, \emph{Mitigation}, \emph{Location}, or \emph{System Failure}. The goal is to discover schema fields that are sufficiently expressive to capture the diversity of historical reports while remaining concise, interpretable, and minimally redundant. %Beyond identifying the fields themselves, an effective schema generation framework should also produce natural-language questions associated with each field, enabling humans to provide the required information in a structured and systematic manner.

%Automatically generating such schemas is particularly challenging because different forms serve fundamentally different operational purposes and therefore require distinct information requirements. For example, some forms emphasize equipment condition and maintenance actions, while others focus on operational impacts, preservation activities, security concerns, or hazard mitigation. Furthermore, domain experts often rely on tacit reporting practices that extend beyond generic requirements and are not fully documented in manuals or reporting guidelines. As a result, manually designing and maintaining structured question sets for every form category is labor-intensive, expensive, and difficult to scale as reporting requirements evolve.

\begin{figure*}[t]
    \centering
    \includegraphics[width=\textwidth]{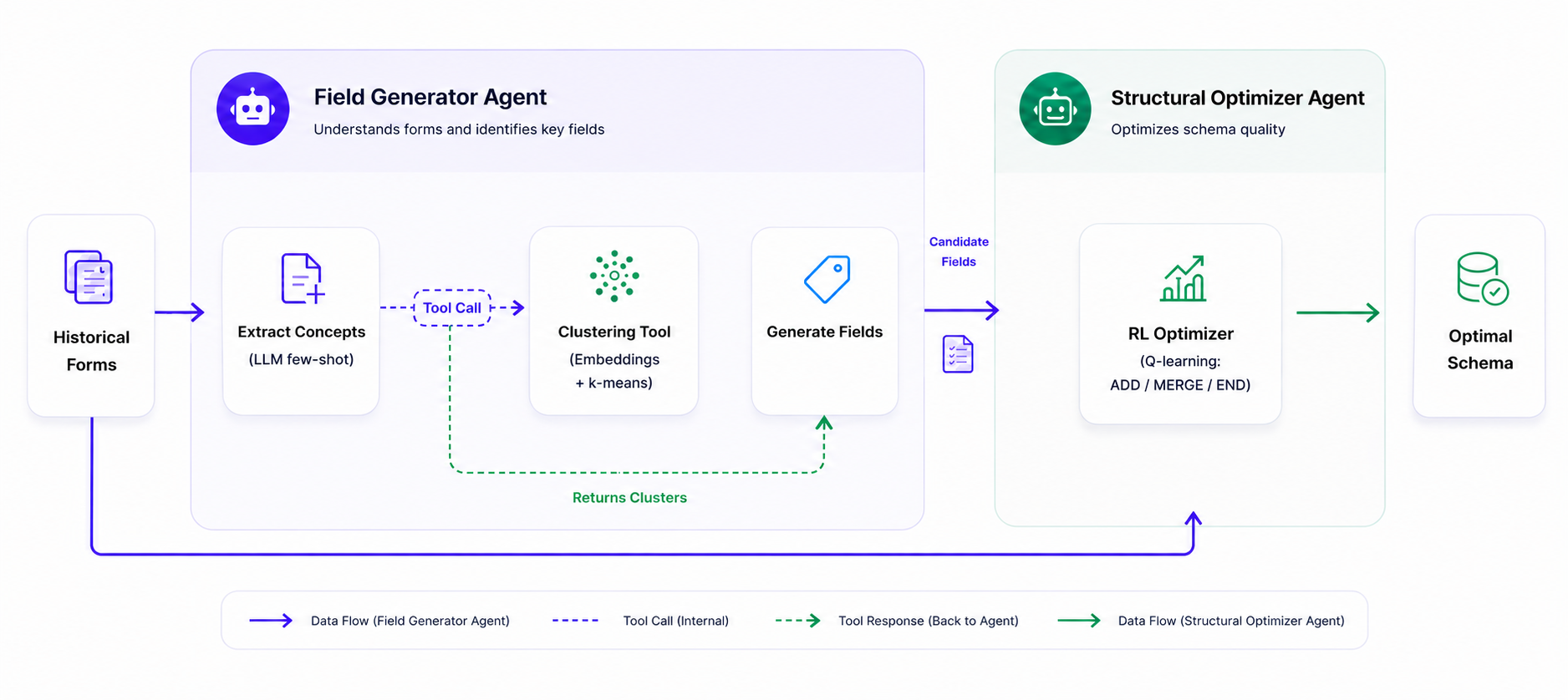}
    \caption{\small Overview of the proposed ASMR framework consisting of a Field Generator Agent for semantic concept extraction and candidate field generation, and a Structural Optimizer Agent for RL-based schema optimization.}
    \label{fig:schema_framework}
\end{figure*}

To address this challenge, we study the automatic schema generation problem and propose {\bf ASMR}, a modular agentic framework that combines Large Language Model (LLM)-based semantic understanding with reinforcement learning-based optimization techniques. The framework consists of two specialized agents, as illustrated in Figure~\ref{fig:schema_framework}. The first, a \emph{Field Generator Agent}, analyzes historical reports, extracts salient semantic concepts from unstructured narratives, adaptively clusters related concepts at multiple granularities, and generates candidate schema fields. The second, a \emph{Structural Optimizer Agent}, models schema construction as a sequential decision-making problem and employs reinforcement learning-based optimization to iteratively retain, merge, or discard candidate fields in order to produce compact, informative, and non-redundant schemas. The key insight is that schema design can be formulated as a multi-objective optimization problem that balances coverage, informativeness, compactness, and redundancy.

{\em \bf ASMR's} ability to automatically generate high-quality schemas has several downstream benefits. First, it enables interactive reporting systems in which personnel are guided through structured questions derived from generated schema, improving report completeness, consistency, and adherence to reporting best practices. Second, the resulting standardized reports support downstream analytics such as predictive maintenance~\cite{white2025computational}, operational readiness assessment, anomaly detection, trend analysis, and knowledge extraction. Third, structured schemas improve interoperability across units, platforms, and reporting systems by reducing ambiguity and enabling more reliable aggregation and comparison of operational data. Finally, because the proposed framework learns schemas directly from historical reports, it can continuously adapt to evolving reporting requirements and emerging operational practices without requiring extensive manual redesign by domain experts.

Although {\em \bf ASMR} is motivated by and evaluated on ship operational and maintenance forms, we believe it has applicability to a broad range of industrial, manufacturing, healthcare, logistics, aviation, and infrastructure management systems that rely heavily on semi-structured and unstructured forms and face similar challenges. Development of {\em \bf ASMR} is currently in progress, and our initial evaluations provide encouraging evidence. Compared to recent systems such as PALIMPZEST~\cite{palim}, ELEET~\cite{eleet}, DocETL~\cite{docetl}, and TabAgent~\cite{wu2025tabagent}, the proposed work addresses a complementary problem centered on automatic schema generation for human-AI collaborative operational reporting.

\smallskip \noindent{\bf Contributions.}
\begin{itemize}
\item We formulate the \emph{automatic schema generation problem} for operational and maintenance reporting and propose {\bf ASMR}, a modular agentic framework that combines LLM-based semantic concept extraction, adaptive multi-granularity clustering, and reinforcement learning-based optimization to automatically generate compact, informative, and non-redundant schemas from historical reports. \hfill \mbox{}

\item We demonstrate how the generated schemas can enable structured human-AI collaboration by automatically producing field-specific guidance and questions that assist report authors in creating more complete, consistent, and actionable operational reports.

\item We develop quantitative metrics for evaluating schema quality and demonstrate through preliminary experiments that {\em ASMR} produces compact, informative, and low-redundancy schemas that support effective AI-assisted operational reporting.
\end{itemize}

The rest of the paper is organized as follows: in Section~\ref{sec:datamodel}, we present our data model and formalize the proposed problem. Section~\ref{sec:asmr} presents the proposed modular agentic framework, including the Field Generator Agent, Structural Optimizer Agent, and schema quality evaluation methodology. Section~\ref{sec:exp} presents preliminary experimental evaluation and results. Related work is discussed in Section~\ref{sec:related}. We discuss challenges and open problems in Section~\ref{sec:open} and conclude in Section~\ref{sec:conc}.

\section{Data Model \& Problem Definition}\label{sec:datamodel}

\noindent \paragraph{Historical Forms.}
Let $\mathcal{H} = \{f_1, f_2, \dots, f_n\}$ denote a collection of historical maintenance forms spanning multiple operational categories. Each form instance $f_i \in \mathcal{H}$ corresponds to a free-form narrative associated with a form type $t$, such as \emph{VOIDS AND COFFERDAMS}, \emph{LIVING QUARTERS COMPARTMENTS}, \emph{STORAGE COMPARTMENTS}, or \emph{FUEL OIL TANKS}. These narratives contain heterogeneous operational observations, maintenance descriptions, hazards, mitigation procedures, and equipment-related information expressed in unstructured natural language.

\noindent \paragraph{Concepts.}
A concept represents a semantically meaningful atomic unit extracted from a historical form narrative. Formally, given a form instance $f_t$, a concept $c \in \mathcal{C}$ corresponds to an identifiable semantic entity, operational descriptor, condition, or event appearing in the narrative. Examples of concepts extracted from the forms include \textit{corrosion, water intrusion, improper sealing, water separator assemblies, degraded lagging,} etc. Concepts are generated by the Field Generator Agent using an LLM-driven semantic extraction mechanism and capture the fine-grained semantics present in historical reports.

\noindent \paragraph{Fields.}
A field represents a structured attribute generated by aggregating semantically related concepts across multiple historical forms. Formally, a field $\phi \in \Phi$ corresponds to a canonical abstraction over a subset of concepts $\mathcal{C}_{\phi} \subseteq \mathcal{C}$ that share operational or semantic similarity. For example, concepts such as \emph{Leakage}, \emph{Leak}, and \emph{System Failure} may map to a failure-related field, while concepts such as \emph{Location}, \emph{Accessibility}, and \emph{Bulkhead} may map to a compartment-related field. The resulting field space $\Phi$ defines the candidate field space generated by the Field Generator Agent and subsequently optimized into schemas for different form types.

\noindent

\paragraph{Schema of a Form Type.}  
Given a form type $t$, its schema $\mathcal{S}_t$ is defined as a structured collection of fields that collectively characterize the semantic and operational information expected to appear in forms of type $t$. Formally,
\[
\mathcal{S}_t = \{\phi_1, \phi_2, \dots, \phi_m\},
\]
where each $\phi_i \in \Phi$ represents a canonical field generated by aggregating semantically related concepts extracted from historical narratives belonging to form type $t$. The schema captures the recurring semantic structure underlying a category of operational reports.

For example, the schema corresponding to the form type \texttt{FUEL OIL TANKS} may contain fields such as
\[
\mathcal{S}_{\texttt{FuelOilTanks}} =
\left\{
\begin{array}{l}
\texttt{Condition}, \\
\texttt{Location}, \\
\texttt{FailureDescription}, \\
\texttt{ElectricalSystems}, \\
\texttt{Mitigation}
\end{array}
\right\}
\]
where the field \texttt{FailureDescription} may aggregate concepts such as \emph{Leakage}, \emph{Leak}, and \emph{System Failure}, while \texttt{ElectricalSystems} may aggregate concepts such as \emph{Electrical} and \emph{Electronics}. Similarly, the schema for \texttt{VOIDS AND COFFERDAMS} may contain fields such as \texttt{Location}, \texttt{Hazard}, \texttt{Condition}, \texttt{Accessibility}, and \texttt{Preservation}.

\noindent \paragraph{\bf Problem Definition.}  
Given a collection of historical form narratives
\[
\mathcal{H} = \{f_1, f_2, \dots, f_n\},
\]
where each form instance $f_i \in \mathcal{H}$ belongs to one of $m$ different form types
\[
\mathcal{T} = \{t_1, t_2, \dots, t_m\},
\]
and each instance consists of unstructured free-form operational text, the objective is to automatically discover an optimal structured schema
\[
\mathcal{S}_{t_j} = \{\phi_1, \phi_2, \dots, \phi_k\}
\]
for every form type $t_j \in \mathcal{T}$ such that the schema captures the recurring semantic structure underlying the historical forms belonging to that form type.

Our proposed agentic framework addresses this problem through two specialized agents:

(i) a \emph{Field Generator Agent} that extracts a set of semantic concepts

\[
\mathcal{C} = \{c_1, c_2, \dots, c_r\}
\]

from historical narratives, groups semantically related concepts to consolidate overlapping information across multiple forms and form types, and generates an overcomplete set of candidate fields

\[
\Phi = \{\phi_1, \phi_2, \dots, \phi_p\},
\]

and

(ii) a \emph{Structural Optimizer Agent} that optimizes the candidate field space to construct the final schema $\mathcal{S}_{t_j}$ for each form type $t_j$.
\section{{\em \bf ASMR}}\label{sec:asmr}

{\em \bf ASMR} is a modular agentic framework for automatic schema generation from historical ship maintenance and operational forms. It mainly consists of two specialized agents: a Field Generator Agent that extracts semantic concepts and generates an overcomplete candidate field space, and a Structural Optimizer Agent that applies reinforcement learning-based optimization to construct compact, informative, and non-redundant schemas for different form types.

Figure~\ref{fig:schema_framework} presents an overview of the overall {\em \bf ASMR} framework. Section~\ref{sec:field_agent} describes the Field Generator Agent, while Section~\ref{sec:optimizer_agent} presents the Structural Optimizer Agent. To quantitatively assess the quality of the generated schemas by ASMR, we further introduce our schema quality evaluation methodology, described in Section~\ref{sec:schema_eval}.

\begin{comment}
\begin{figure}[t]
\centering
\includegraphics[width=\columnwidth]{figures/ASMR-Stages.png}
\caption{Overview of the 4 key stages of the {\em \bf ASMR} framework: 1) semantic concept extraction, 2) candidate field generation, 3) RL-based structural optimization, and 4) schema quality evaluation.}
\label{fig:asmrframework}
\end{figure}    
\end{comment}

\vspace{-10pt}

\subsection{Field Generator Agent}\label{sec:field_agent}

The Field Generator Agent is an LLM-based agent that discovers and abstracts semantic information requirements from historical maintenance narratives. As illustrated in Figure~\ref{fig:field_generator_agent}, the agent extracts semantic concepts from unstructured reports, leverages a clustering tool to organize related concepts into coherent groups, and generates an overcomplete candidate field space. The resulting candidate fields are then provided to the Structural Optimizer Agent for schema optimization.

\subsubsection{Semantic Concept Extraction}

Historical forms are first grouped according to their corresponding form types using ESWBS codes (Engineering Systems Breakdown Structure identifiers used to categorize operational and maintenance components). For each form type $t \in \mathcal{T}$, representative subsets of historical forms are sampled from $\mathcal{H}$ and provided to the LLM as the core semantic engine of the Field Generator Agent.

The concept extraction process relies on instruction-guided few-shot prompt engineering, where sampled historical forms and task-specific parameters, such as the number of extracted concepts, are incorporated into the prompt. This enables flexible control over the scope and granularity of the extracted semantic concepts.

The extracted outputs correspond to atomic semantic concepts rather than final schema fields. Example concepts extracted from \texttt{FUEL OIL TANKS} forms include:

\begin{quote}
\small
\texttt{non-destructive test, berthing compartment, tank top, blast, warped seating surface}
\end{quote}

These concepts form the semantic foundation for the subsequent clustering and field abstraction stages.

\subsubsection{Clustering Tool Invocation}

To organize extracted concepts into higher-level semantic groupings, the Field Generator Agent invokes a clustering tool over concept embeddings. For each form type $t \in \mathcal{T}$, the extracted concepts are embedded into a semantic vector space and clustered using K-means across multiple values of $K$ to capture semantic abstractions at different granularities. The agent progressively explores larger values of $K$ until it determines that additional refinements no longer yield meaningful semantic abstractions. The resulting concept groups are then aggregated to form an overcomplete set of candidate semantic groups for downstream field abstraction.

For example, $cluster_1=\{\texttt{Ripped}, \texttt{Cracked}, \texttt{Loose}\}$ and $cluster_2=\{\texttt{Fall risk}, \texttt{Trip hazard}\}$ may form two semantically coherent clusters representing related operational characteristics. This clustering stage consolidates noisy and overlapping concepts extracted from heterogeneous maintenance narratives.

\subsubsection{Field Abstraction}

After generating concept groups, the Field Generator Agent abstracts each cluster into a higher-level schema candidate field representing the shared semantic meaning of the concepts within that cluster. The abstraction prompt incorporates both clustered concept instances and representative historical forms, enabling the agent to leverage the operational context in which the concepts appear when generating candidate fields.

For example, $cluster_1=\{\texttt{Ripped}, \texttt{Cracked}, \texttt{Loose}\}$ may be abstracted as \texttt{Condition}, while $cluster_2=\{\texttt{Fall risk}, \texttt{Trip hazard}\}$ may be abstracted as \texttt{Hazard}.

The union of generated field labels across different concept groups and clustering granularities constitutes the overcomplete candidate field space $\Phi$. Table~\ref{tab:generatedfields} presents examples of generated candidate fields across multiple ship operational form categories. The resulting candidate field space is subsequently provided to the Structural Optimizer Agent for schema optimization.

\begin{table*}[t]
\centering
\small
\begin{tabular}{p{2.8in}p{4.3in}}
\toprule
\textbf{Form Type} & \textbf{Initial Fields} \\
\midrule
VOIDS AND COFFERDAMS & Location, Condition, Hazard, Preservation, Accessibility \\

COMPARTMENTS NOT OTHERWISE COVERED & Condition, Location, Hazard, Mitigation, System, Maintenance, Security, Safety, Bulkhead \\

STORAGE COMPARTMENTS (incl. ammo, chem, hazmat) & Condition, System Piping, Mitigation, Location, Equipment, Hazard \\

FUEL OIL TANKS & Leakage, Condition, Mitigation, Location, Leak, Electrical, Electronics \\

WASTE TANKS & Condition, Location, System, Pumps, Equipment, Piping, System Piping, Tank Condition, System Failure \\
\bottomrule
\end{tabular}
\vspace{-0.4em}
\caption{Examples of generated candidate fields across multiple ship operational form categories.}
\label{tab:generatedfields}
\end{table*}

\begin{figure}[t]
\centering
\includegraphics[width=\linewidth]{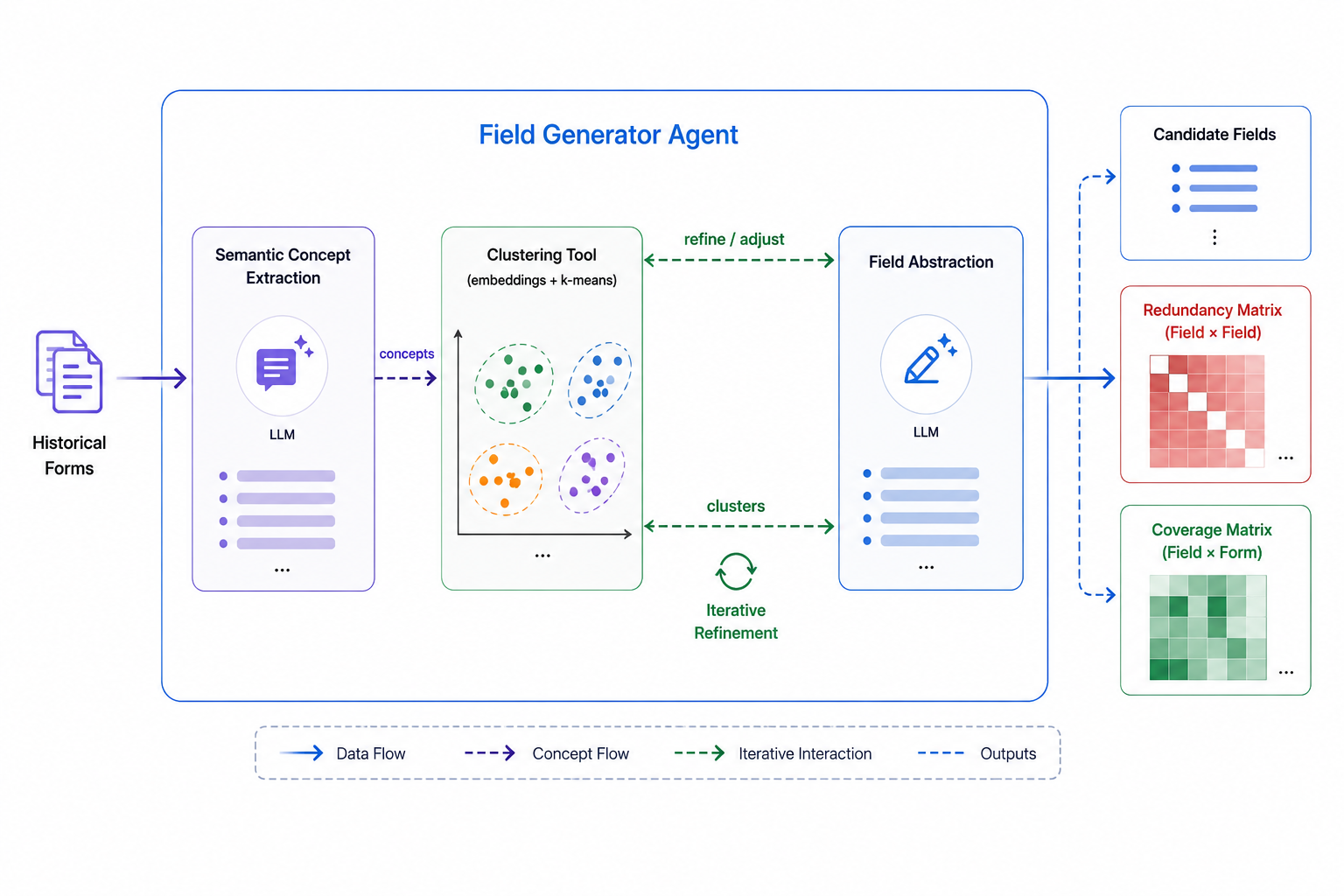}
\caption{Example snapshot of the Field Generator Agent. The agent extracts semantic concepts and generates candidate schema fields for downstream schema optimization.}
\label{fig:field_generator_agent}
\end{figure}

\subsection{Structural Optimizer Agent}\label{sec:optimizer_agent}

After the Field Generator Agent produces the overcomplete candidate field space $\Phi$, the resulting candidate fields are communicated to the Structural Optimizer Agent, which is responsible for constructing compact, informative, and non-redundant schemas for each form type. The agent formulates schema optimization as a stochastic Markov Decision Process (MDP) and sequentially refines the candidate field space through field selection, field merging, and schema termination decisions. As shown previously in Figure~\ref{fig:field_generator_agent}, the Field Generator Agent produces not only candidate schema fields but also the field--field redundancy matrix and field--form coverage matrix, all of which are utilized throughout the RL-based schema optimization process described next.

\subsubsection{RL Formulation}

\paragraph{State Space}

A state is defined as $s = \langle \phi, t \rangle$
where $\phi \in \Phi$ denotes the current candidate schema field and $t \in \mathcal{T}$ denotes the corresponding form type.

Additional states include:

\begin{equation*}
\langle START, t \rangle, \quad \langle END, t \rangle
\end{equation*}

where $\langle START, t \rangle$ initializes schema construction for form type $t$, while $\langle END, t \rangle$ terminates the current schema generation episode.

\paragraph{Action Space}

The action space consists of:

\begin{equation*}
\mathcal{A} = \{ADD, MERGE, STOP\}
\end{equation*}

where:

\begin{itemize}[leftmargin=*]
    \item $ADD$ adds the next candidate field to the current schema,
    \item $MERGE$ merges the current field with the next candidate field,
    \item $STOP$ transitions the agent to the terminal $\langle END, t \rangle$ state.
\end{itemize}

\subsubsection{Environment Transition Modeling}

\begin{figure}[t]
\centering
\includegraphics[width=0.55\textwidth]{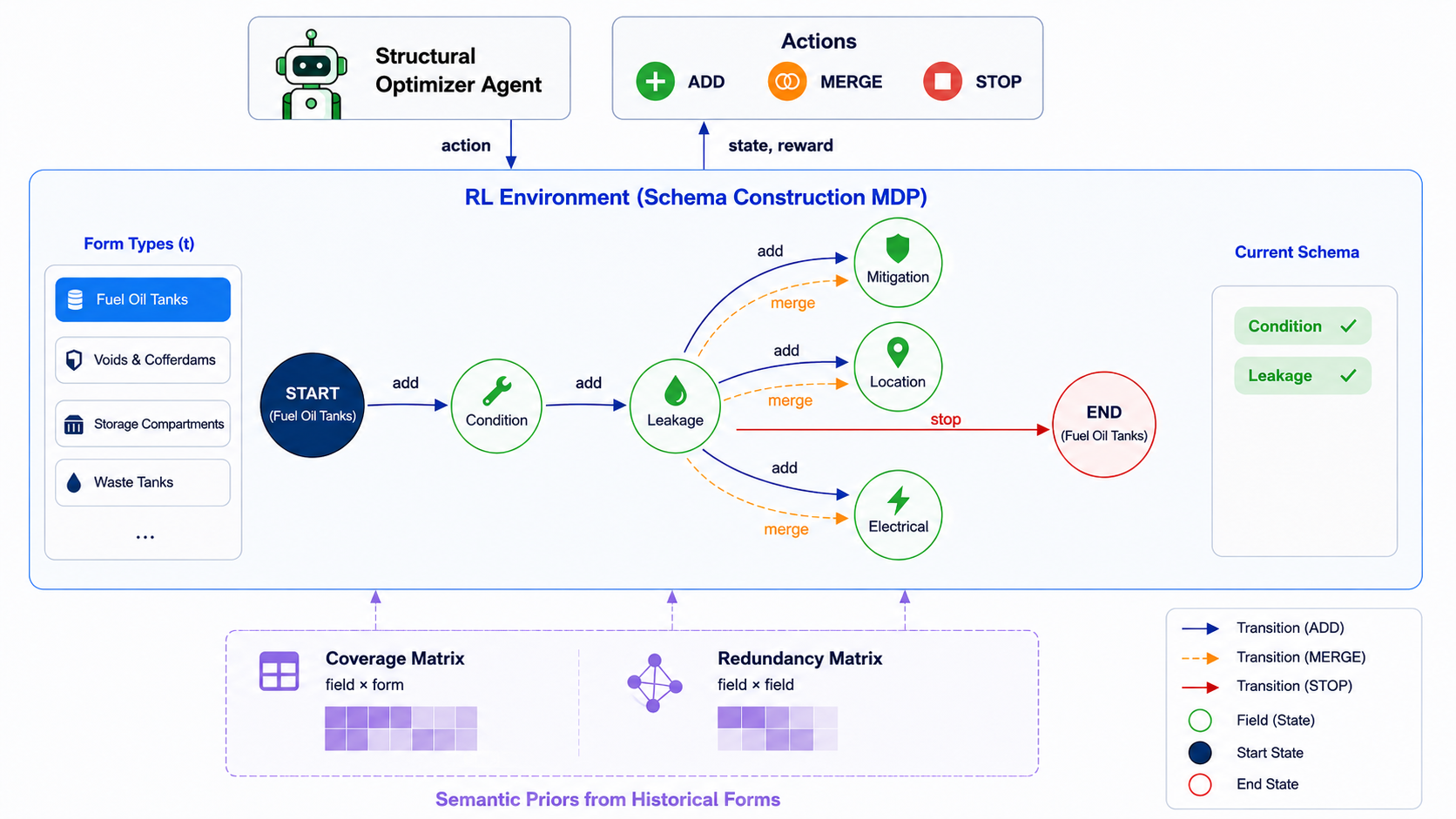}
\caption{Example snapshot of the Structural Optimizer Agent operating in a stochastic RL environment for schema optimization.}
\label{fig:rlsnapshot}
\end{figure}

We model the RL environment as a stochastic environment whose transition probabilities are guided by semantic priors extracted from historical operational forms. Figure~\ref{fig:rlsnapshot} illustrates a snapshot of the schema construction process.

Given a current state $\langle \phi_k, t \rangle$, an $ADD$ action transitions the agent toward another candidate field $\langle \phi_j, t \rangle$, with probability proportional to the occurrence frequency of $\phi_j$ within historical forms of type $t$:

\begin{align}
P_{add}(\phi_j,t)
\propto
freq(\phi_j,t) + noise_{add},
\quad \forall \phi_j \in \Phi,\ \phi_j \neq \phi_k
\end{align}

Field frequencies are derived from a binary field--form coverage matrix $C$ generated by the Field Generator Agent:

\begin{equation}
C(\phi_i, f_j)=
\begin{cases}
1 & \text{if field } \phi_i \text{ appears in form } f_j \\
0 & \text{otherwise}
\end{cases}
\end{equation}

where field presence is determined through LLM-guided semantic verification of the concepts associated with $\phi_i$.

Similarly, a $MERGE$ action transitions according to precomputed semantic redundancy statistics:

\begin{align}
P_{merge}(\phi_j)
\propto
M(\phi_j,\phi_k,t) + noise_{merge},
\quad \forall \phi_j \in \Phi,\ \phi_j \neq \phi_k
\end{align}

where $M$ denotes a field--field semantic redundancy matrix with entries

\begin{equation}
M(\phi_i,\phi_j) \in [0,1].
\end{equation}

Redundancy scores are computed using LLM-assisted semantic comparisons over the concepts associated with each field. Small stochastic noise terms are included to encourage exploration during RL training.

Both field frequency statistics and redundancy scores are precomputed prior to optimization, separating expensive semantic computations from the online RL process. During each episode, $ADD$ and $MERGE$ actions continuously update the current schema representation used for reward computation.

\subsubsection{Reward Function}

The reward function jointly optimizes semantic coverage, schema compactness, and redundancy minimization. Given a transition from state $s_t$ to $s_{t+1}$ under action $a$, the reward is defined as:

\begin{equation}
\label{reward_signal}
R(s_t,s_{t+1},a)
=
\alpha \cdot \Delta Coverage
-
\beta \cdot \Delta Size
-
\gamma \cdot \Delta Redundancy
\end{equation}

where
\vspace{-1pt}
\begin{align}
\Delta Coverage
&=
Coverage(S_{t+1}) - Coverage(S_t)
\\
\Delta Size
&=
Size(S_{t+1}) - Size(S_t)
\\
\Delta Redundancy
&=
Redundancy(S_{t+1}) - Redundancy(S_t)
\end{align}

and $S_t$ and $S_{t+1}$ denote the schema before and after the transition, respectively.

Since fields may be merged during optimization, each schema element corresponds to a field group $g_i \subseteq \Phi$ containing one or more candidate fields.

Coverage is computed using the field--form coverage matrix $C$:

\begin{equation}
Coverage(f_j,g_i)
=
\frac{1}{|g_i|}
\sum_{\phi_k \in g_i}
C(\phi_k,f_j)
\end{equation}

\begin{equation}
Coverage(S)
=
\sum_{f_j \in \mathcal{H}}
\sum_{g_i \in S}
Coverage(f_j,g_i)
\end{equation}

Similarly, redundancy is computed using the field--field semantic redundancy matrix $M$:

\begin{equation}
Redundancy(g_i,g_j)
=
\frac{1}{|g_i||g_j|}
\sum_{\phi_a \in g_i}
\sum_{\phi_b \in g_j}
M(\phi_a,\phi_b)
\end{equation}

\begin{equation}
Redundancy(S)
=
\sum_{g_i \in S}
\sum_{g_j \in S,\ j>i}
Redundancy(g_i,g_j)
\end{equation}

And schema compactness is measured as $Size(S)=|S|$.

These quantities are efficiently recomputed after each action and incorporated into the reward signal in Equation~\ref{reward_signal} during RL optimization.

\begin{comment}
   \subsubsection{RL Optimization}

We use Temporal Difference Learning (TDL)-based Q-learning for schema optimization. Prior to RL training, the framework preprocesses historical operational forms to compute the field-form coverage matrix $C$ and the field-field semantic redundancy matrix $M$ used throughout environment transition modeling and reward computation.

The Structural Optimizer Agent subsequently learns the Q-table over stochastic schema construction transitions through repeated training episodes. The Q-values are updated according to the standard Temporal Difference learning formulation:

\begin{equation}
Q(s_t,a_t)
\leftarrow
Q(s_t,a_t)
+
\alpha
\Big[
r_t
+
\gamma
\max_{a'}
Q(s_{t+1},a')
-
Q(s_t,a_t)
\Big]
\label{eq:qlearning}
\end{equation}

where $\alpha$ denotes the learning rate, $\gamma$ denotes the discount factor, and $r_t$ corresponds to the reward computed using Equation~\ref{reward_signal}.

During inference, the agent starts from $\langle START, t \rangle$ and iteratively follows the learned policy until reaching the terminal $\langle END, t \rangle$ state, producing the final optimized schema for form type $t$.

\begin{figure*}[t]
\centering
\includegraphics[width=0.95\textwidth]{figures/RL_Snapshot.png}
\caption{Example snapshot of the Structural Optimizer Agent operating in a stochastic RL environment for schema optimization.}
\label{fig:rlsnapshot}
\end{figure*} 
\end{comment}

\subsubsection{RL Optimization}

We use Temporal Difference Learning (TDL)-based Q-learning for schema optimization. Prior to training, the field--form coverage matrix $C$ and field--field semantic redundancy matrix $M$ are precomputed and used throughout environment transition modeling and reward computation.

The Structural Optimizer Agent learns a Q-table over stochastic schema construction transitions. Q-values are updated according to:

\begin{equation}
Q(s_t,a_t)
\leftarrow
Q(s_t,a_t)
+
\alpha
\Big[
r_t
+
\gamma
\max_{a'}
Q(s_{t+1},a')
-
Q(s_t,a_t)
\Big]
\label{eq:qlearning}
\end{equation}

where $\alpha$ is the learning rate, $\gamma$ is the discount factor, and $r_t$ is the reward from Equation~\ref{reward_signal}.

During inference, the agent starts from $\langle START, t \rangle$ and follows the learned policy until reaching $\langle END, t \rangle$, producing the optimized schema for form type $t$.

\paragraph{Schema Finalization.}

After optimization converges, the Structural Optimizer Agent assigns each finalized field group a human-readable label and guide description. This process uses an LLM that receives the merged field groups, representative historical forms, and pairwise semantic statistics. The resulting schema consists of the finalized field structure together with field-level guidance text supporting human-AI collaboration for form completion during inference.

\vspace{-3pt}

\paragraph{Schema Population Inference.}

Once generated, the finalized schema can be applied to unseen operational narratives through a schema population inference process. During inference, an LLM-based Field-Value Extraction Tool leverages the schema fields and their guide descriptions to map unstructured narratives into the generated schema. For each field, the tool extracts a corresponding value when present and returns an associated confidence score.

\paragraph{Human-AI Collaboration.}

The extracted values and confidence scores can be incorporated into a human-AI collaborative workflow during schema population at inference time. Beyond confidence-aware verification, generated field descriptions can provide field-specific guidance and help generate clarification questions for report authors. Fields with low confidence or missing information may trigger targeted questions that help users provide additional details, improving report completeness and consistency while focusing human effort on uncertain schema components.
\subsection{Schema Quality Evaluation}
\label{sec:schema_eval}

We evaluate schemas generated by {\em \bf ASMR} using six dataset-driven metrics designed to assess coverage, support, consistency, informativeness, redundancy, and compactness. Table~\ref{tab:schema_metrics} summarizes the corresponding formulations.

Coverage measures how well schema fields represent historical forms, while support captures the probability ($P$) of observing a schema pattern across the dataset. Consistency measures conditional associations among schema components, and informativeness is estimated using lift-based statistics to quantify non-random relationships between fields. Redundancy quantifies semantic overlap using the precomputed redundancy matrix $M$, while schema size measures overall schema compactness. Together, these metrics provide a quantitative framework for evaluating the coverage, coherence, compactness, and representational utility of generated schemas.

\vspace{-3pt}

\begin{table}[H]
\centering
\small
\setlength{\tabcolsep}{6pt}
\renewcommand{\arraystretch}{1.05}

\begin{tabular}{p{1.1in}p{2.1in}}
\toprule
\textbf{Metric} & \textbf{Formula} \\
\midrule

Coverage &
$\displaystyle
=
\frac{1}{|\mathcal{H}||S|}
\sum_{f_j \in \mathcal{H}}
\sum_{g_i \in S}
Coverage(f_j,g_i)
$
\\[0.2ex]

Support &
$\displaystyle
= P(S)
$
\\[0.2ex]

Consistency &
$\displaystyle
=
\frac{1}{|S|(|S|-1)}
\sum_{g_i \in S}
\sum_{g_j \in S,\; j \neq i}
P(g_j|g_i)
$
\\[0.2ex]

Informativeness &
$\displaystyle
=
\frac{1}{|S|(|S|-1)}
\sum_{g_i \in S}
\sum_{g_j \in S,\; j \neq i}
Lift(g_i,g_j)
$
\\[0.2ex]

Redundancy &
$\displaystyle
=
\frac{1}{|S|(|S|-1)}
\sum_{g_i \in S}
\sum_{g_j \in S,\; j \neq i}
Redundancy(g_i,g_j)
$
\\[0.2ex]

Schema Size &
$\displaystyle
= |S|
$
\\

\bottomrule
\end{tabular}
\vspace{-0.5em}
\caption{Schema quality evaluation metrics.}
\label{tab:schema_metrics}
\end{table}

\vspace{-10pt}
\section{Preliminary Experimental Results}\label{sec:exp}

We conduct a preliminary experimental evaluation of {\em \bf ASMR} on multiple categories of ship maintenance and operational forms in order to assess the effectiveness of the proposed schema generation and optimization framework.

\subsection{Experimental Setup}

Experiments were conducted on multiple ship maintenance and operational form categories, including several representative form types presented in Table~\ref{tab:generatedfields}, each containing approximately 500 historical forms.

The Field Generator Agent used GPT-4o Mini for semantic concept extraction and field abstraction, together with embedding-based semantic clustering as an external tool. The Structural Optimizer Agent used TDL-based Q-learning with the stochastic transition model and reward formulation described in Section~\ref{sec:asmr}. Hyperparameters, including the learning rate, discount factor, reward weights, and clustering granularity, were selected via grid search.

Algorithms were implemented in Python 3.11 and executed on an HPC cluster consisting of 6 nodes with 2.45 GHz AMD EPYC 7753 processors and 512 GB RAM.

\begin{table}[t]
\centering
\small
\begin{tabular}{lc}
\toprule
\textbf{Component} & \textbf{Running Time} \\
\midrule

Field Generator: Concept Extraction & $\approx$20 min \\
Field Generator: Clustering Tool & $\approx$5 min \\
Field Generator: Field Abstraction & $\approx$10 min \\
Field Generator: Coverage \& Redundancy Statistics & $\approx$35 min \\

\midrule

Structural Optimizer: RL Training & $\approx$8 sec \\
Structural Optimizer: Schema Finalization & $\approx$2 sec \\

\midrule

Inference: Field-Value Extraction Tool & $\approx$2 sec \\

\bottomrule
\end{tabular}
\caption{Observed running times of major components within the proposed framework.}
\label{tab:runtime}
\end{table}

The Field Generator Agent dominates the overall computational cost due to LLM-assisted concept extraction, field abstraction, and construction of the coverage and redundancy statistics required for downstream optimization. In contrast, the Structural Optimizer Agent remains computationally lightweight since the required semantic statistics and transition structures are precomputed prior to optimization. Furthermore, inference remains highly efficient, requiring only a few seconds per form.

\subsection{Quantitative Results}

\subsubsection{Metric-Based Evaluation Results}

We evaluate the generated schemas using the dataset-driven evaluation metrics introduced in Section~\ref{sec:schema_eval}, including coverage, support, consistency, informativeness, redundancy, and schema size. We compare three progressively refined schema representations:

\begin{itemize}[leftmargin=*]
    \item {\bf Raw Concepts}: atomic semantic concepts directly extracted from historical forms,

    \item {\bf Candidate Schema}: the overcomplete candidate field space generated by the Field Generator Agent,

    \item {\bf Optimized Schema (ASMR)}: the final schema generated by the Structural Optimizer Agent.
\end{itemize}

Table~\ref{tab:metricresults} summarizes the average metric values across multiple operational form categories.

\begin{table*}[t]
\centering
\small
\begin{tabular}{lcccccc}
\toprule
\textbf{Schema Representation} &
\textbf{Coverage} &
\textbf{Support} &
\textbf{Consistency} &
\textbf{Informativeness} &
\textbf{Redundancy} &
\textbf{Schema Size} \\
\midrule

Raw Concepts & 0.19 & $<0.01$ & 0.17 & 0.94 & 0.67 & 42.0 \\

Candidate Schema & 0.48 & 0.08 & 0.74 & 1.43 & 0.39 & 8.3 \\

Optimized Schema (ASMR) & \textbf{0.64} & \textbf{0.21} & 0.69 & \textbf{1.81} & \textbf{0.17} & \textbf{5.4} \\

\bottomrule
\end{tabular}
\caption{Average quantitative evaluation results across multiple operational form categories.}
\label{tab:metricresults}
\vspace{-15pt}
\end{table*}

The results demonstrate a clear progression in schema quality across the three schema representations. Raw semantic concepts produce highly fragmented and redundant schemas with low statistical support, limited coverage, and weak semantic consistency. After processing by the Field Generator Agent, the candidate schema representation substantially improves coverage, support, consistency, and informativeness while significantly reducing schema redundancy and overall schema size.

The final schemas generated by the Structural Optimizer Agent achieve the strongest overall performance across most evaluation metrics. In particular, the Structural Optimizer Agent significantly improves schema compactness and reduces semantic redundancy while further increasing schema coverage and semantic informativeness. Although the consistency metric slightly decreases after RL optimization, the resulting schemas exhibit stronger non-random semantic associations and more informative schema components overall. These results demonstrate that the proposed agentic framework can successfully combine semantic field generation and schema optimization to identify compact and semantically meaningful schema representations from heterogeneous operational narratives.

\subsubsection{Inference Statistics}

To further evaluate the practical usability of the generated schemas during deployment, we analyze inference-time schema population statistics over unseen historical operational forms. During inference, the Field-Value Extraction Tool attempts to populate values for each finalized schema field from incoming operational narratives.

For each form type, the Field-Value Extraction Tool evaluates whether a schema field can be successfully populated from a given form instance together with an associated confidence score for the extracted value. The framework then computes the average non-null presence percentage and average confidence score for each schema field across all forms belonging to the corresponding operational category.

\begin{figure}[t]
\centering
\begin{subfigure}[t]{0.48\columnwidth}
\centering
\includegraphics[width=\linewidth]{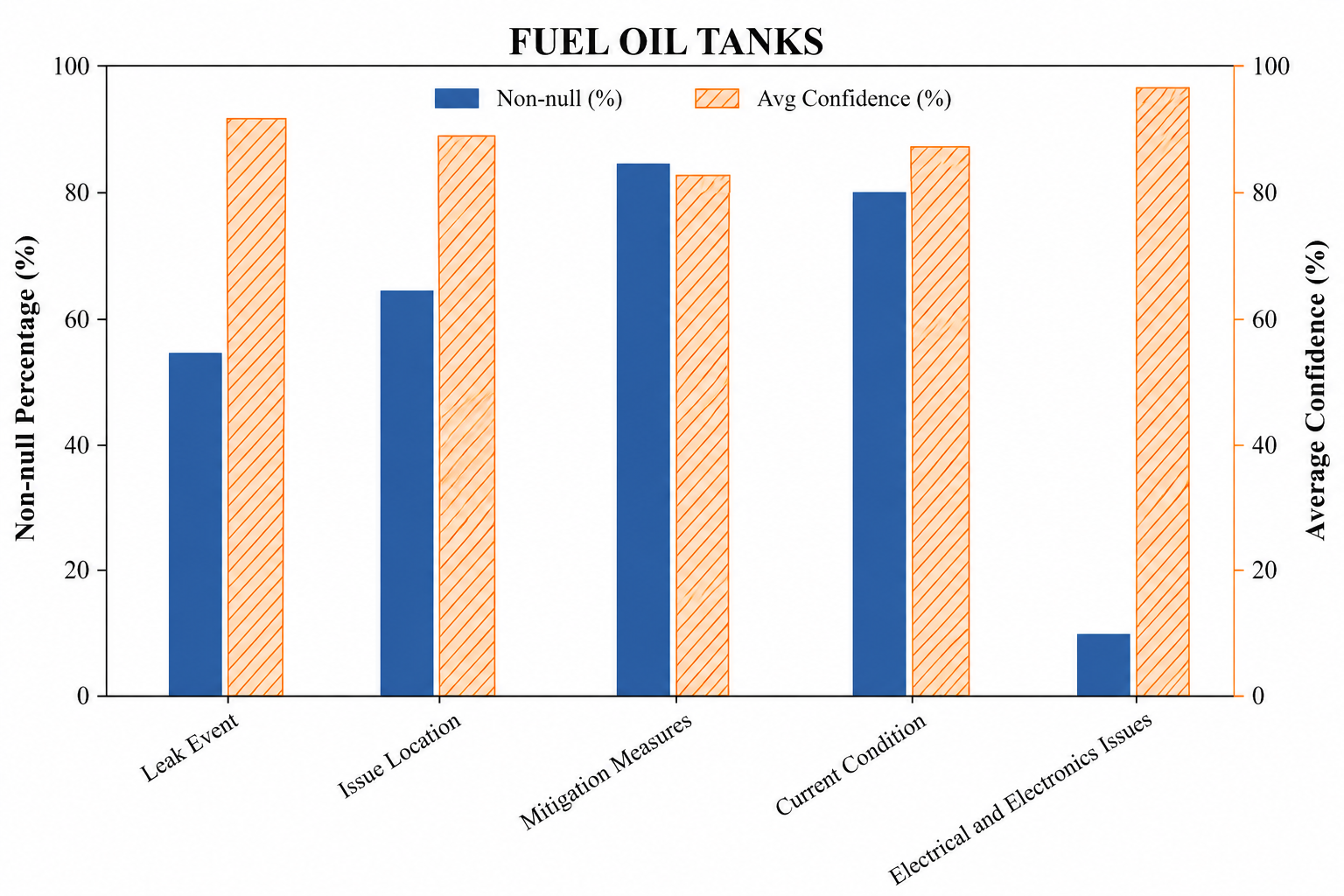}
\caption{\texttt{FUEL OIL TANKS}}
\label{fig:inferencepresence_fuel}
\end{subfigure}
\hfill
\begin{subfigure}[t]{0.5\columnwidth}
\centering
\includegraphics[width=\linewidth]{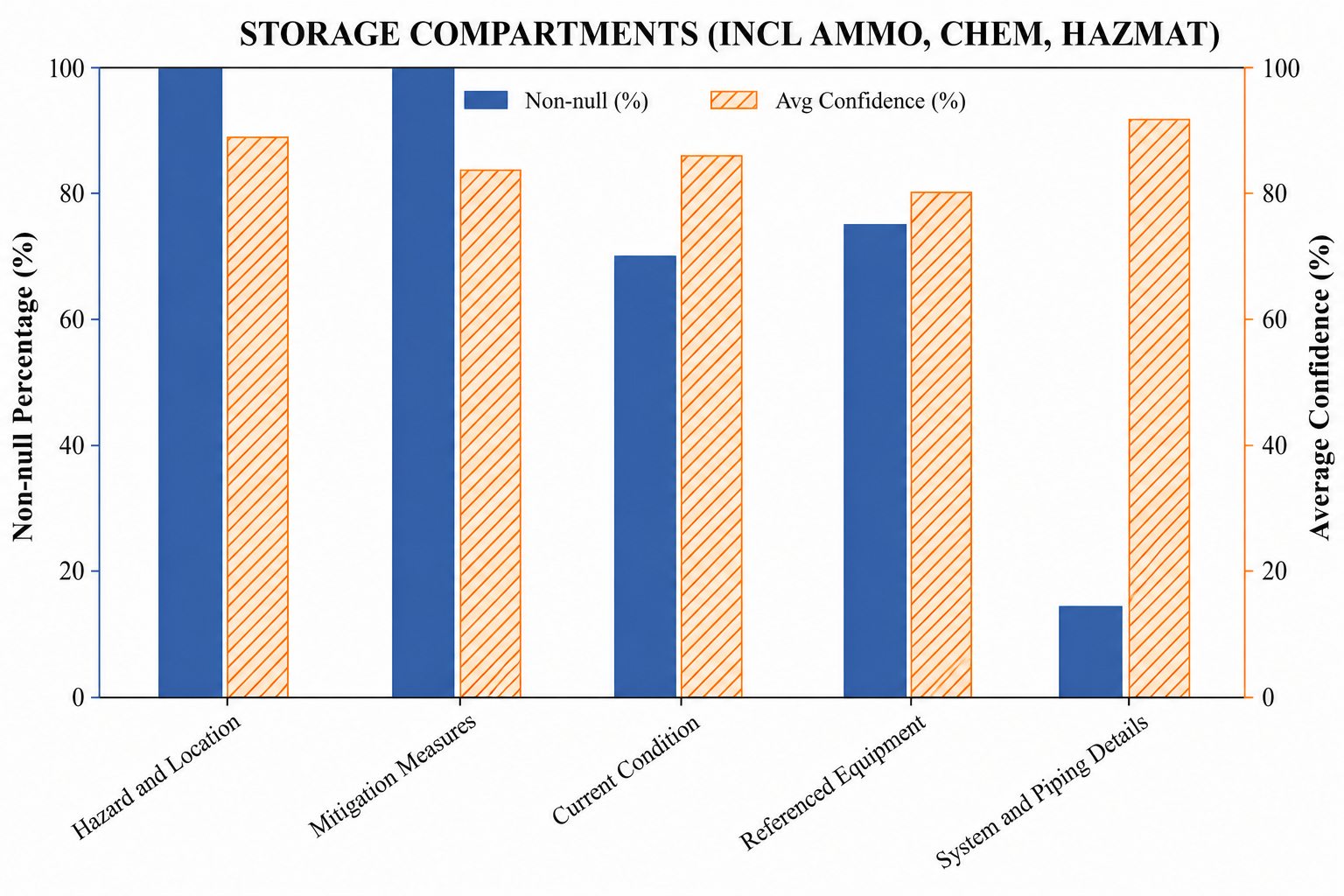}
\caption{\texttt{STORAGE COMPARTMENTS}}
\label{fig:inferencepresence_storage}
\end{subfigure}

\caption{Inference-time field population statistics for finalized schema fields across two operational form categories.}
\label{fig:inferencepresence}
\end{figure}

Figure~\ref{fig:inferencepresence} presents an example inference-time analysis for two representative operational form categories, showing both the percentage of non-null extracted values and the average extraction confidence for each finalized schema field.

The results indicate that operationally important schema fields such as mitigation measures, issue locations, and current conditions achieve both high population frequency and strong extraction confidence during inference. In contrast, highly specialized fields such as electrical and electronics issues exhibit lower population frequencies despite maintaining high confidence when present.

These inference statistics provide useful signals for downstream human-AI collaborative operational workflows. In particular, fields with consistently high presence and confidence can be prioritized during AI-assisted form completion, while low-frequency or uncertain fields may trigger targeted guidance, follow-up questions, or user verification to improve report completeness and consistency.

\subsection{Qualitative Results}

In addition to the quantitative evaluation metrics presented earlier, we further analyze the generated schemas from a qualitative operational perspective in order to better understand how the Structural Optimizer Agent consolidates semantically related fields into compact and interpretable schema representations.

For the \texttt{Compartments Not Otherwise Covered} operational form category, the Structural Optimizer Agent identified that the fields:
\[
Hazard + Safety + Security + System
\]
capture substantially overlapping operational semantics across historical forms. Consequently, the agent merged these fields into a more compact schema representation while preserving the underlying operational meaning. This example further illustrates that the Structural Optimizer Agent can move beyond simple pairwise merging when necessary and progressively consolidate multiple semantically related operational fields into a unified and meaningful schema field.

Similarly, for the \texttt{Waste Tanks} operational form category, the Structural Optimizer Agent identified strong semantic redundancy between fields related to piping and system-level operational descriptions throughout the forms. As a result, fields such as:
\begin{equation*}
Piping + System\_Piping
\end{equation*}
were consolidated into a unified schema component, while redundant operational attributes including overlapping pump and system references were discarded during optimization.

Overall, these qualitative examples demonstrate that the proposed agentic framework not only improves schema quality quantitatively, but also produces compact operational schemas that remain semantically interpretable and operationally meaningful from a human qualitative perspective.

\section{Related Work}\label{sec:related}

Our work falls into the intersection of automatic schema generation, agentic systems for data management, and the evaluation of unstructured data processing pipelines.

$\textbf{Automatic Schema Generation.}$ Automatic schema generation is a relevant problem in the context of many systems. Large Language Models and NLP techniques have been used for schema generation within many areas such as relational databases, document-oriented systems, and data warehousing \cite{db_schema_llm, 10.1007/978-3-030-59003-1_10, 11313182}. However, unlike these structured contexts, the free-form nature of ship maintenance report blocks introduces new challenges. Similar AI-driven methods have been developed for schema discovery in JSON documents, knowledge graphs, and ontologies \cite{Neubauer_2025, zhang2025schemagenerationlargeknowledge, cheung2026generativeontologystructuredknowledge}, demonstrating both the usefulness of LLMs for schema generation and the contextual improvement schemas can provide for text generation, especially in domain-specific environments \cite{du-etal-2020-schema, wang2023grammarpromptingdomainspecificlanguage, oswald2024largelanguagemodelsplanning}. Reinforcement learning \cite{barto2018reinforcement, kaelbling1996reinforcement} has also been proposed for JSON schema generation \cite{lu2025learninggeneratestructuredoutput}, for fine-tuning rather than schema construction. LLM-driven schema generation thus demonstrates the capability to support structured extraction of maintenance knowledge from historical ship reports.

$\textbf{Agentic Data Systems.}$ The proposed work also relates to agentic unstructured data processing systems \cite{eleet, docetl, palim, Pappula_Rusum_2023}, as historical ship maintenance reports are largely composed of unstructured text data. With current LLM capabilities, much work has been done in the analysis of unstructured text and optimization of querying and processing of unstructured data. For example, PALIMPZEST~\cite{palim} proposes a pipeline for the optimization of analytical queries over unstructured data. ELEET~\cite{eleet} focuses on efficient query execution over text and tables using learned multi-modal operators. DocETL~\cite{docetl} proposes agentic optimization and rewriting for complex document processing workflows. DITTO ~\cite{Li_2020} proposes an entity-matching system using pre-trained language models and optimization techniques for domain-specific matching. Our work focuses on a complementary problem, by using unstructured domain-specific data for schema discovery using agents for concept and entity recognition, schema construction, and optimization.

$\textbf{Unstructured Data Pipeline Evaluation.}$ Finally, our work requires a robust evaluation system for determining the quality of constructed schemas. Many schema generation methods for structured data such as in data warehouses \cite{11313182} employ use of ground-truth or manually designed schemas. In contrast, it is more difficult to attain ground-truth schemas for ship maintenance form types due to lack of standardization. Other methods such as the use of validation agents or NLP techniques are also used for operating on unstructured text data and pipelines \cite{docetl, palim}. However, use of these methods is highly task-specific and does not directly transfer to our schema evaluation setting. Instead, we construct a quantitative evaluation framework for maintenance report schemas based on core qualities such as coverage, informativeness, compactness, and non-redundancy. Nevertheless, developing more robust methods to address such limitations remains an important direction for future work.

\section{Challenges and Open Problems}\label{sec:open}

\paragraph{\textbf{Confidence-aware Schema Generation.}}
Large Language Models (LLMs) serve as a key backbone of {\bf ASMR}, supporting semantic concept extraction, redundancy estimation, and candidate field generation. However, LLMs are inherently probabilistic and may produce inconsistent or semantically ambiguous outputs. An important open problem is the development of confidence-aware schema generation frameworks that explicitly model uncertainty in AI-generated schema components. One possible direction is to leverage self-consistency prompting and estimate confidence scores over generated concepts and schema fields. Incorporating uncertainty requires jointly reasoning about semantic quality, redundancy, coverage, and confidence, making uncertainty-aware optimization a significant challenge.

\vspace{-2pt}

\paragraph{\textbf{Reward Function Design and Quantification.}}
Designing an effective reward function remains a central challenge in {\bf ASMR}. The current framework balances semantic coverage and redundancy minimization, which often exhibit an inherent tension: increasing schema fields may improve coverage but introduce overlap, while aggressively reducing redundancy may remove operationally important distinctions. Furthermore, redundancy scores are estimated using LLM-based semantic reasoning, which remains subjective and model-dependent. Developing principled formulations for redundancy and other reward dimensions remains an important research problem.

\vspace{-2pt}
\paragraph{\textbf{Evaluation.}}
A key challenge faced by {\bf ASMR} is the absence of reliable ground truth in historical operational datasets. Reports are often noisy, incomplete, and heterogeneous, making semantic interpretation difficult and subjective. While domain experts can assist with validation, manual curation is costly and difficult to scale. Consequently, evaluation is challenging due to both the lack of benchmark datasets and the absence of a definitive ``correct'' schema for a given form category. Developing objective evaluation criteria and principled validation methodologies therefore remains an important open problem.
\vspace{-2pt}
\paragraph{\textbf{Historical Data Limitations.}}
{\bf ASMR} learns report schemas from historical data, leveraging patterns in past reports to identify information fields for future reporting. However, this approach faces a fundamental bootstrapping challenge: historical data is often incomplete or erroneous due to the very reporting deficiencies that ASMR is intended to address. Consequently, schema-assisted report writing may perpetuate commonly unrecorded information categories. Developing methods to address such limitations, such as data sampling techniques or incorporation of external guidance, remains an important direction for future work.
\section{Conclusion}\label{sec:conc}
This paper presented {\em \bf ASMR}, an agentic framework for generating structured schemas for AI-assisted ship operational and maintenance report writing from historical forms. By combining a Field Generator Agent and a Structural Optimizer Agent, the framework automatically identifies compact and semantically meaningful schemas for heterogeneous operational forms. Initial results demonstrate reduced redundancy while preserving semantic coverage and representational quality. Beyond ship reporting workflows, {\em \bf ASMR} has potential applicability to industrial, healthcare, logistics, aviation, and infrastructure systems relying on large-scale operational documentation. More broadly, this work opens research directions at the intersection of data management, agentic AI, and human-AI collaboration.

\begin{acks}
This work was sponsored by the Office of Naval Research under Contract No. N0001425C2403
\end{acks}

\bibliographystyle{ACM-Reference-Format}
\bibliography{sample}

@String{Computing = "Computing" }

@String{Computer = "{IEEE} Computer" }

@String{Springer = "Springer-Verlag" }

@article{palim,
  title={A declarative system for optimizing ai workloads},
  author={Liu, Chunwei and Russo, Matthew and Cafarella, Michael and Cao, Lei and Chen, Peter Baille and Chen, Zui and Franklin, Michael and Kraska, Tim and Madden, Samuel and Vitagliano, Gerardo},
  journal={arXiv preprint arXiv:2405.14696},
  year={2024}
}

@article{eleet,
  title={Efficient learned query execution over text and tables [technical report]},
  author={Urban, Matthias and Binnig, Carsten},
  journal={arXiv preprint arXiv:2410.22522},
  year={2024}
}

@article{docetl,
  title={Docetl: Agentic query rewriting and evaluation for complex document processing},
  author={Shankar, Shreya and Chambers, Tristan and Shah, Tarak and Parameswaran, Aditya G and Wu, Eugene},
  journal={arXiv preprint arXiv:2410.12189},
  year={2024}
}

@inproceedings{white2025computational,
  title={A Computational Framework for Estimating Days of Maintenance Delay of Naval Ships.},
  author={White, Gerald and Mistry, Deep and Chhoa, Kevin and Roy, Senjuti Basu and Zhang, Lingyi and Bienkowski, Adam and Pattipati, Krishna R},
  booktitle={EDBT},
  pages={1014--1022},
  year={2025}
}

@article{db_schema_llm,
author = {Salem, Nadia and Al-Tarawneh, Khawla and Hudaib, Amjad and Salem, Hamza and Tareef, Afaf and Salloum, Hadi and Mazzara, Manuel},
year = {2024},
month = {12},
pages = {1703-1713},
title = {Generating database schema from requirement specification based on natural language processing and large language model},
volume = {16},
journal = {Computer Research and Modeling},
doi = {10.20537/2076-7633-2024-16-7-1703-1713}
}

@InProceedings{10.1007/978-3-030-59003-1_10,
author="G{\'o}mez, Paola
and Casallas, Rubby
and Roncancio, Claudia",
editor="Hartmann, Sven
and K{\"u}ng, Josef
and Kotsis, Gabriele
and Tjoa, A. Min
and Khalil, Ismail",
title="Automatic Schema Generation for Document-Oriented Systems",
booktitle="Database and Expert Systems Applications",
year="2020",
publisher="Springer International Publishing",
address="Cham",
pages="152--163",
isbn="978-3-030-59003-1"
}

@INPROCEEDINGS{11313182,
  author={Abdelrahman, Abdelrahman A. and Elbahrawy, Abdelrahman A. and Sobieh, Ahmed R. and ElSaid, Alaa E. and Ali, Ahmed M. and Elsharawy, Arwa A. and Shaaban, Yasmine and Afify, Yasmine M.},
  booktitle={2025 Twelfth International Conference on Intelligent Computing and Information Systems (ICICIS)}, 
  title={DataForge: An AI-Driven Data Warehouse Schema Generator}, 
  year={2025},
  volume={},
  number={},
  pages={677-684},
  doi={10.1109/ICICIS66182.2025.11313182}}

@inproceedings{Neubauer_2025,
   title={AI-assisted JSON Schema Creation and Mapping},
   url={http://dx.doi.org/10.1109/MODELS-C68889.2025.00019},
   DOI={10.1109/models-c68889.2025.00019},
   booktitle={2025 ACM/IEEE 28th International Conference on Model Driven Engineering Languages and Systems Companion (MODELS-C)},
   publisher={IEEE},
   author={Neubauer, Felix and Uekermann, Benjamin and Pleiss, Jürgen},
   year={2025},
   month=Oct, pages={79–83} }

@misc{zhang2025schemagenerationlargeknowledge,
      title={Schema Generation for Large Knowledge Graphs Using Large Language Models}, 
      author={Bohui Zhang and Yuan He and Lydia Pintscher and Albert Meroño Peñuela and Elena Simperl},
      year={2025},
      eprint={2506.04512},
      archivePrefix={arXiv},
      primaryClass={cs.AI},
      url={https://arxiv.org/abs/2506.04512}, 
}

@misc{cheung2026generativeontologystructuredknowledge,
      title={Generative Ontology: When Structured Knowledge Learns to Create}, 
      author={Benny Cheung},
      year={2026},
      eprint={2602.05636},
      archivePrefix={arXiv},
      primaryClass={cs.AI},
      url={https://arxiv.org/abs/2602.05636}, 
}

@inproceedings{du-etal-2020-schema,
    title = "Schema-Guided Natural Language Generation",
    author = "Du, Yuheng  and
      Oraby, Shereen  and
      Perera, Vittorio  and
      Shen, Minmin  and
      Narayan-Chen, Anjali  and
      Chung, Tagyoung  and
      Venkatesh, Anushree  and
      Hakkani-Tur, Dilek",
    editor = "Davis, Brian  and
      Graham, Yvette  and
      Kelleher, John  and
      Sripada, Yaji",
    booktitle = "Proceedings of the 13th International Conference on Natural Language Generation",
    month = dec,
    year = "2020",
    address = "Dublin, Ireland",
    publisher = "Association for Computational Linguistics",
    url = "https://aclanthology.org/2020.inlg-1.35/",
    doi = "10.18653/v1/2020.inlg-1.35",
    pages = "283--295"
}

@misc{wang2023grammarpromptingdomainspecificlanguage,
      title={Grammar Prompting for Domain-Specific Language Generation with Large Language Models}, 
      author={Bailin Wang and Zi Wang and Xuezhi Wang and Yuan Cao and Rif A. Saurous and Yoon Kim},
      year={2023},
      eprint={2305.19234},
      archivePrefix={arXiv},
      primaryClass={cs.CL},
      url={https://arxiv.org/abs/2305.19234}, 
}

@misc{oswald2024largelanguagemodelsplanning,
      title={Large Language Models as Planning Domain Generators}, 
      author={James Oswald and Kavitha Srinivas and Harsha Kokel and Junkyu Lee and Michael Katz and Shirin Sohrabi},
      year={2024},
      eprint={2405.06650},
      archivePrefix={arXiv},
      primaryClass={cs.CL},
      url={https://arxiv.org/abs/2405.06650}, 
}

@article{Pappula_Rusum_2023, title={Multi-Modal AI for Structured Data Extraction from Documents}, volume={4}, url={https://ijeret.org/index.php/ijeret/article/view/273}, DOI={10.63282/3050-922X.IJERET-V4I3P109}, abstractNote={Structured data extraction of unstructured documents like scanned pictures, PDF documents, or photos has become a crucial task to accomplish in a wide range of industries in a world that is becoming more and more digitalized. In the following paper, we present a multi-modal artificial intelligence system combining the visual layout analysis with the capability of natural language processing (NLP) to extract structured fields of heterogeneous documents. The offered solution would use convolutional neural networks (CNNs) and transformer-based models to group the interpretation of the spatial layouts, textual contexts, and semantics in a combined manner. The system has proved to be resistant to document formatting inconsistencies, noise, skew, and complex typography by integrating these features. The hybrid architecture initially carries out visual parsing and identifies regions of interest and yields hierarchical layout features. Such features are combined with semantic embeddings trained on pre-trained NLP models like BERT or LayoutLM, allowing the context-aware extraction of fields. The model is trained and tested on the various types of documents in three domains, including insurance claims, billing statements and legal contracts. The performance metrics depict a considerable increase in punctuality and recollected accuracy compared to conventional OCR-based guideline schemes and multimodal one-dimensional models. This study shows the impact of cross-modal reasoning style to resolve the typical obstacles of lacking labels, ambiguous fields, and varying arrangements. The modular structure of the system is also domain-adaptable and extensible, which paves the way for scalable and automated document understanding in enterprise solutions
}, number={3}, journal={International Journal of Emerging Research in Engineering and Technology}, author={Pappula, Kiran Kumar and Rusum, Guru Pramod}, year={2023}, month={Oct.}, pages={75–86} }

@misc{lu2025learninggeneratestructuredoutput,
      title={Learning to Generate Structured Output with Schema Reinforcement Learning}, 
      author={Yaxi Lu and Haolun Li and Xin Cong and Zhong Zhang and Yesai Wu and Yankai Lin and Zhiyuan Liu and Fangming Liu and Maosong Sun},
      year={2025},
      eprint={2502.18878},
      archivePrefix={arXiv},
      primaryClass={cs.CL},
      url={https://arxiv.org/abs/2502.18878}, 
}

@article{Li_2020,
   title={Deep entity matching with pre-trained language models},
   volume={14},
   ISSN={2150-8097},
   url={http://dx.doi.org/10.14778/3421424.3421431},
   DOI={10.14778/3421424.3421431},
   number={1},
   journal={Proceedings of the VLDB Endowment},
   publisher={Association for Computing Machinery (ACM)},
   author={Li, Yuliang and Li, Jinfeng and Suhara, Yoshihiko and Doan, AnHai and Tan, Wang-Chiew},
   year={2020},
   month=Sept, pages={50–60} }

@book{barto2018reinforcement,
  title={Reinforcement learning: An introduction},
  author={Barto, Andrew G and Sutton, Richard S},
  year={2018}
}

@article{kaelbling1996reinforcement,
  title={Reinforcement learning: A survey},
  author={Kaelbling, Leslie Pack and Littman, Michael L and Moore, Andrew W},
  journal={Journal of artificial intelligence research},
  volume={4},
  pages={237--285},
  year={1996}
}

@inproceedings{wu2025tabagent,
  title={TabAgent: A Multi-Agent Table Extraction Framework for Unstructured Documents},
  author={Wu, Jingfei and Han, Junyi and Gao, Yujin},
  booktitle={2025 5th International Symposium on Artificial Intelligence and Big Data (AIBDF)},
  pages={600--607},
  year={2025},
  organization={IEEE}
}

\end{document}